\pdfoutput=1
\documentclass[11pt]{article}

\usepackage[preprint]{acl}

\usepackage{times}
\usepackage{latexsym}
\usepackage{booktabs}
\usepackage{amsmath}
\usepackage{amssymb}

\usepackage[T1]{fontenc}
\usepackage[utf8]{inputenc}

\usepackage{microtype}

\usepackage{inconsolata}

\usepackage{graphicx}
\usepackage{courier}

\newcommand{\uzhlogo}{\includegraphics[width=1.3em]{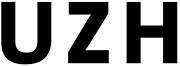}}
\newcommand{\zhawlogo}{\includegraphics[width=1em]{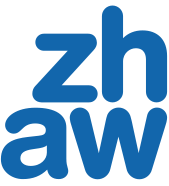}}

\title{Audio Description Generation in the Era of LLMs and VLMs: \\ A Review of Transferable Generative AI Technologies}

\author{$\textbf{Yingqiang Gao}^{\uzhlogo}$, 
$\textbf{Lukas Fischer}^{\uzhlogo}$,
$\textbf{Alexa Lintner}^{\zhawlogo}$, $\textbf{Sarah Ebling}^{\uzhlogo}$
\vspace{0.8em}
\\
$^{\uzhlogo}\text{Department of Computational Linguistics, University of Zurich, Switzerland}$ \\ 
\texttt{\{yingqiang.gao, fischerl, ebling\}@cl.uzh.ch} \\
$^{\zhawlogo}\text{School of Applied Linguistics, Zurich University of Applied Sciences, Switzerland}$ \\
\texttt{alexa.lintner@zhaw.ch} \\
 }

\definecolor{urlpink}{RGB}{255, 105, 180} 
\definecolor{citeblue}{RGB}{0, 102, 204}

\hypersetup{
    colorlinks=true,      
    linkcolor=urlpink,        
    filecolor=magenta,    
    urlcolor=urlpink,         % Color of external URLs
    linkbordercolor=red,   % Border color around links when not using colorlinks=true
}

\begin{document}
\maketitle
\begin{abstract}
% Audio descriptions (ADs) function as acoustic commentaries designed to assist blind persons and persons with visual impairments in comprehending digital  media content on television and in movies, among other settings. 
% As an accessibility service typically provided by trained AD professionals, the generation of ADs demands significant human effort, making the process both time-consuming and cost-inefficient. 
% Recent advancements in natural language processing (NLP) and computer vision (CV), particularly in large language models (LLMs) and vision-language models (VLMs), have paved the way for the full automation of AD generation. 
% This survey reviews the cutting-edge technologies pertinent to AD generation in the era of LLMs and VLMs: we aim to discuss how state-of-the-art NLP and CV technologies can be applied to generate ADs and to identify essential research directions for the future.  

Audio descriptions (ADs) function as acoustic commentaries designed to assist blind persons and persons with visual impairments in accessing digital  media content on television and in movies, among other settings. 
As an accessibility service typically provided by trained AD professionals, the generation of ADs demands significant human effort, making the process both time-consuming and costly. 
Recent advancements in natural language processing (NLP) and computer vision (CV), particularly in large language models (LLMs) and vision-language models (VLMs), have allowed for getting a step closer to automatic AD generation. 
This paper reviews the technologies pertinent to AD generation in the era of LLMs and VLMs: we discuss how state-of-the-art NLP and CV technologies can be applied to generate ADs and identify essential research directions for the future.

\end{abstract}

\section{Introduction}

\subsection{Background}

The formalization of AD as a public service can be traced back only to the early 1980s in the United States \citep{mazur2020audio}. Initially introduced in the theater, AD services have expanded to a wide range of contexts, including television programs, movies, art galleries, and museums, in order to mitigate the information loss experienced by blind individuals and individuals with visual impairments. A significant milestone in the development of ADs was achieved in 2010, when the European Parliament included the provision of ``accessible audiovisual media services'' in its directives for that year \citep{reviers2016audio}. Since then, AD research has garnered widespread interest.

\begin{figure}
    \centering
    \includegraphics[width=\columnwidth]{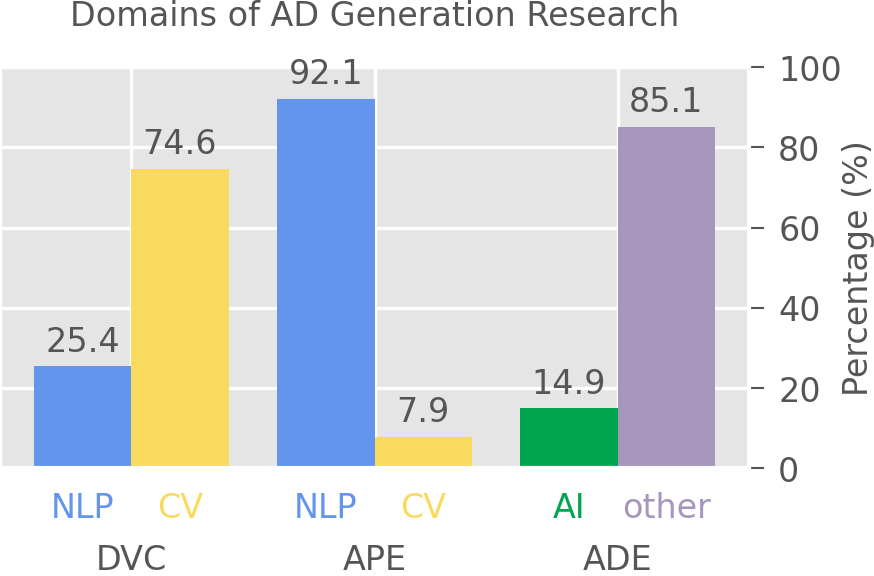}
    \caption{Domain contributions of AD-generation-related publications reviewed in this survey. DVC, APE, and ADE represent the three main steps of AD generation systems: Dense Video Captioning, AD Post-Editing, and Audio Description Evaluation, respectively. The figure illustrates the varying contributions to AD generation research across different domains. For ADE, ``other'' represents non-AI-related research disciplines such as psychology.}
    \label{fig:teaser}
\end{figure}

\begin{figure*}
    \centering
    \includegraphics[width=\textwidth]{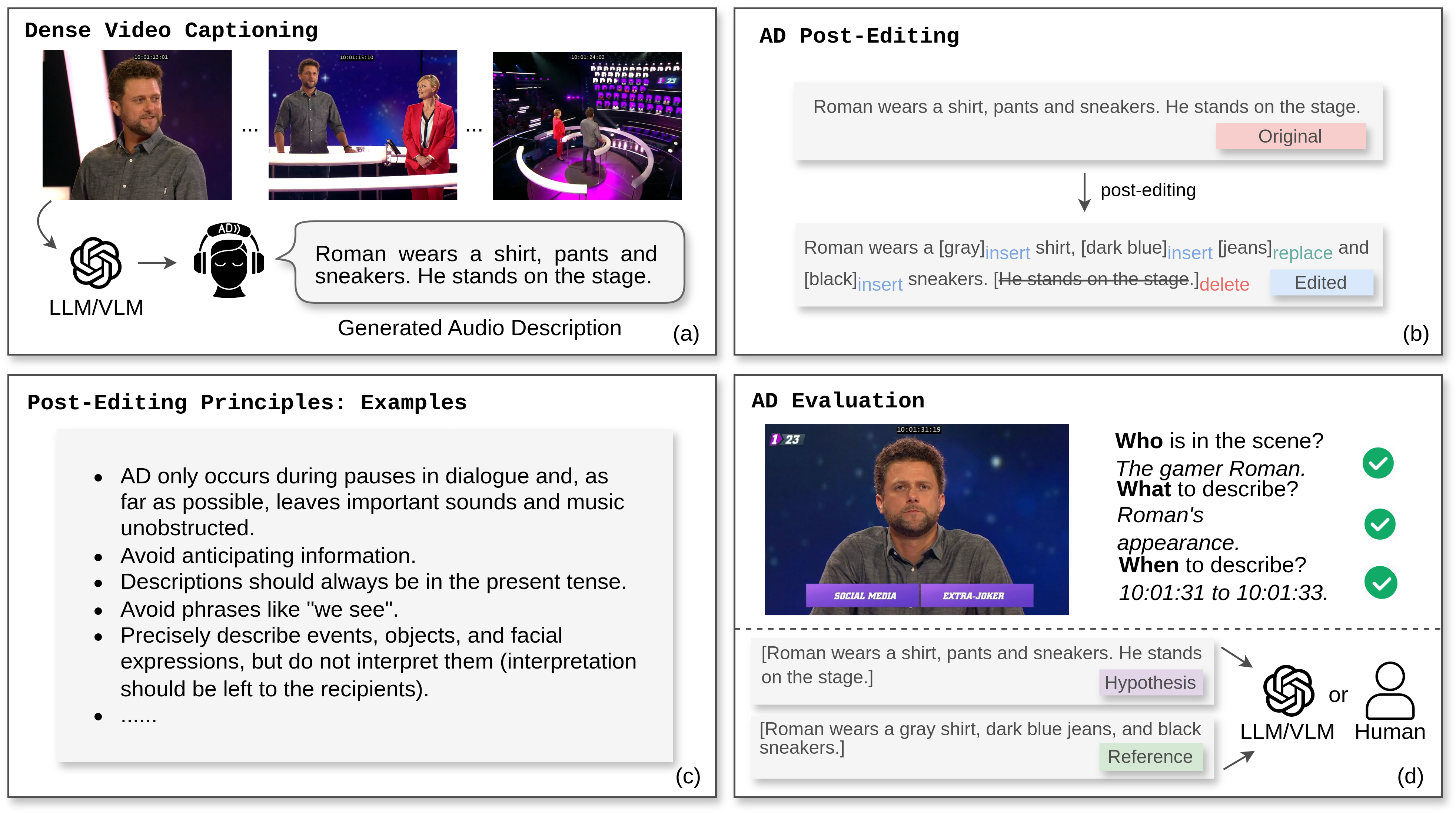}
    \caption{Components of modern AD generation systems with LLM/VLM participation. \textbf{(a) Dense Video Captioning}: the task of generating AD scripts from the given video clips; \textbf{(b) AD Post-Editing}: the task of polishing the generated AD scripts; \textbf{(c) AD Script Creation Guidelines}: used as guidance for Post-Editing; \textbf{(d) AD Evaluation}: Quality assessment of generated ADs. Example taken from the Swiss TV show \textit{1 gegen 100}.}
    \label{fig:pipeline}
\end{figure*}

ADs are traditionally produced by professional audio describers. The production process begins with acquiring the broadcast material, ideally complemented with time codes. Audio describers then review the material and create AD scripts (ideally in cooperation with blind audio describers or audio describers with visual impairments) tailored to the broadcast content. The final step involves recording the AD scripts in a studio, potentially with the involvement of a blind prooflistener, or synthesizing the speech and subsequently mixing the acoustic ADs with the original broadcast audio \citep{fryer2016introduction}. Producing ADs for a 90-minute movie can take approximately 35 to 40 working hours\footnote{Experience shared by audio describers we work with.}, underscoring the vast amount of information that remains inaccessible to blind and visually impaired individuals without these ADs. This highlights the significant value and necessity of the work carried out by professional audio describers.

However, training professional audio describers is a time-intensive process \citep{matamala2007designing, jankowska2017blended, colmenero2019training, mazur2021audio, yan2023audio}. In addition, depending on the provider, AD scripts are sometimes required in multiple languages.  Consequently, a shortage of qualified describers exists, leading to high service costs and unmet demand for accessible media.

\subsection{Motivation}

As the demand for AD generation continues to grow due to reinforced legal requirements \citep{braun2022automating}, both the NLP and CV community have dedicated efforts to solve this problem. 
In recent years, generative AI technologies such as LLMs \citep{brown2020language, touvron2023llama} and VLMs \citep{radford2021learning, ramesh2022hierarchical, li2023videochat, zhang2023video} have demonstrated remarkable capabilities in addressing numerous real-world challenges, including text and image generation. These advancements pave the way for the (semi-)automation of AD generation, as the crucial steps of generating ADs can be offloaded to these large models with significantly less human involvement. Figure~\ref{fig:pipeline} depicts the components of a typical modern AD generation system with three crucial steps:

\paragraph{Dense Video Captioning (DVC)} Given a video, the task is to generate AD scripts that consist of informative descriptions. This inherently multimodal task requires integrating both visual and textual features to create coherent and contextually appropriate AD scripts.

\paragraph{AD Post-Editing (APE)} After generating the initial ADs, refine them according to a set of pre-defined principles. This post-editing process ensures that the ADs meet specific quality standards and accurately convey the intended information. Note that given sufficient performance of the preceding DVC step, this step would not be necessary; however, the state-of-the-art is such that the APE stage is not yet dispensable.

\paragraph{AD Evaluation (ADE)} The generated ADs must undergo both quantitative and qualitative assessments, ideally with the involvement of the target groups. This evaluation process measures the effectiveness, accuracy, and overall quality of the ADs, ensuring they meet the necessary criteria for accessibility and usability.

In this survey, we investigate   generative AI technologies for developing AD generation systems, with a \textbf{special focus on the participation of LLMs/VLMs}. Specifically, we concentrate on the latest research outcomes in NLP and CV (i.e. papers published from 2020 onward, signifying the release of GPT-3 by OpenAI).

\definecolor{dataset}{rgb}{0.99, 0.76, 0.0}
\definecolor{method}{rgb}{0.19, 0.55, 0.91}
\definecolor{mix}{rgb}{0.0, 0.65, 0.31}

\begin{table*}[!htb]
    \centering
    \resizebox{\textwidth}{!}{
    \begin{tabular}{l|lccc|l}
    \toprule
    Research & Venue & Task & Video Encoder & Text Decoder & Method | Dataset \\ 
    \midrule
    %\citet{dinh2024trafficvlm} & CVPR'24 & VFE & CLIP-ViT-L/14 & T5-Base & \textcolor{method}{TrafficVLM} \\
    %\citet{islam2024video} & CVPR'24 & VFE & TimeSformer + DistilBERT & GPT-2(3.5)/FLAN-T5 & \textcolor{mix}{Video ReCap-(U) | Ego4D-HCap} \\
    %\citet{kim2024you} & CVPR'24 & VFE & CLIP-ViT-L/14 & Deformable Transformer & \textcolor{method}{CM$^2$} \\
    \citet{yun2024shvit} & CVPR'24 & VFE & Vanilla ViT & not applicable & \textcolor{method}{SHViT} \\
    \citet{hassani2023neighborhood}  & CVPR'23 & VFE & Swin Transformer & not applicable & \textcolor{method}{NAT} \\
    \citet{chen2023efficient} & ICCV'23 & VFE & Vanilla ViT & not applicable & \textcolor{method}{EVAD} \\
    \citet{liu2023efficientvit} & CVPR'23 & VFE & EfficientViT & not applicable & \textcolor{method}{EfficientViT} \\
    \citet{zhao2022tuber} & CVPR'22 & VFE & Vanilla ViT & not applicable & \textcolor{method}{TubeR} \\
    \citet{liu2022video} & CVPR'22 & VFE & Swin Transformer & not applicable & \textcolor{method}{Video Swin Transformer} \\
    \citet{yang2022lite} & CVPR'22 & VFE & Vanilla ViT & not applicable & \textcolor{method}{Lite Vision Transformer} \\
    \citet{wu2022pale} & AAAI'22 & VFE & Vanilla VFE & not applicable & \textcolor{method}{Pale Transformer} \\
    \citet{korbar2022personalised} & BMVC'22 & VFE & CLIP-ViT-B/32 & not applicable & \textcolor{mix}{CLIP-PAD | CiA} \\
    \citet{yin2022vit} & CVPR'22 & VFE & DeiT & not applicable & \textcolor{method}{A-ViT} \\ 
    \citet{wu2022memvit} & CVPR'22 & VFE & MViT-v2 & not applicable & \textcolor{method}{MeMViT} \\
   
    \citet{brown2021face} & ICCV'21 & VFE & ResNet50 & not applicable & \textcolor{mix}{MuHPC | VPCD} \\
    \citet{huang2021shuffle} & arXiv'21 & VFE & Swin Transformer & not applicable & \textcolor{method}{Shuffle Transformer} \\
    \citet{liu2021swin} & ICCV'21 & VFE & Hierarchical ViT & not applicable & \textcolor{method}{Swin Transformer} \\
    \citet{rao2021dynamicvit} & NeurIPS'21 & VFE & ViT/DeiT/LV-ViT & not applicable & \textcolor{method}{DynamicViT} \\ 
    \citet{wu2020context} & ECCV'20 & VFE & ResNet50-FPN & not applicable & \textcolor{method}{Context-Aware RCNN} \\
    \citet{huang2020movienet} & ECCV'20 & VFE & not applicable & not applicable & \textcolor{dataset}{MovieNet} \\
    \citet{kukleva2020learning} & CVPR'20 & VFE & NesNeXt-101 & not applicable & \textcolor{method}{LIReC} \\ 
    
    \midrule
    Research & Venue & Task & Video Encoder & Text Decoder & Method | Dataset \\ 
    \midrule
    \citet{chu2024llm} & arXiv'24 & DCG & GPT-4V & GPT-4 & \textcolor{method}{LLM-AD} \\
    \citet{he2024ma} & CVPR'24 & DCG & CLIP-ViT-G/14 + Q-Former & Vicuna-v1.5 & \textcolor{method}{MA-LLM} \\
    \citet{luo2024shotluck} & arXiv'24 & DCG & TinyLlaVA (SigLIP) & TinyLlaVA (TinyLlama/Phi-2) & \textcolor{method}{Shotluck Holmes} \\
    \citet{maaz2024videogpt+} & arXiv'24 & DCG & CLIP-ViT-L/14 + InternVideo-v2 & Vicuna-v1.5/LlaMA-3/Phi3-Mini & \textcolor{mix}{VideoGPT+ | VCGBench-Diverse} \\
    \citet{ye2024mmad} & COLING'24 & DCG & Video-LlaVA-v0 & LlaMA-2 & \textcolor{method}{MMAD} \\
    \citet{yue2024movie101v2} & arXiv'24 & DCG & VideoChat-2/Qwen-VL & GPT-4V & \textcolor{dataset}{Movie101v2-(zh/en)} \\
    \citet{Zhou_2024_CVPR} & CVPR'24 & DCG & CLIP-ViT-L/14 & T5-Base & \textcolor{method}{Streaming DVC} \\
    \citet{blanco2024live} & arXiv'24 & DCG & Deformable Transformer & Deformable Transformer & \textcolor{method}{LVC} \\
    \citet{xie2024autoad} & arXiv'24 & DCG & VideoLlaMA-(2) & LlaMA-3/Gemma-2 & \textcolor{mix}{AutoAD-Zero | TV-AD} \\
    \citet{han2024autoadiii} & CVPR'24 & DCG & Q-Former & OPT/LlaMA-2 & \textcolor{mix}{AutoAD III | (CMD/HowTo)-AD} \\
    \citet{yue2023movie101} & ACL'23 & DCG & Transformer & Transformer & \textcolor{mix}{MNScore | Movie101-{zh}} \\
    \citet{jung2023retrieval} & ACL'23 & DCG & LXMERT & EMT + PDVC & \textcolor{method}{KOFCL} \\
    \citet{han2023autoadii} & ICCV'23 & DCG & CLIP-ViT-B/32 & GPT-2 & \textcolor{mix}{AutoAD II | MAD-(t-eval/L-char)} \\
    \citet{han2023autoad} & CVPR'23 & DCG & CLIP-ViT-B/32 & GPT-2 & \textcolor{mix}{AutoAD | MAD-v2/AudioVault} \\
    \citet{shen2023accurate} & ICCV'23 & DCG & CLIP-ViT-L/14 & Multimodal Transformer & \textcolor{method}{CoCap} \\
    \citet{yang2023multicapclip} & ACL'23 & DCG & CLIP-ViT-B/16 & Vanilla Transformer & \textcolor{method}{MultiCapClip} \\
    \citet{lin2023mm} & arXiv'23 & DCG & GPT-4V & GPT-4 & \textcolor{method}{MM-VID} \\
    \citet{han2023shot2story20k} & arXiv'23 & DCG & CLIP-ViT-L/14 + Q-Former & MiniGPT-4/GPT-4 & \textcolor{dataset}{Shot2Story20K} \\
    \citet{soldan2022mad} & CVPR'22 & DCG & CLIP-ViT-B/32 & not applicable  & \textcolor{dataset}{MAD} \\
    \citet{zhang2022movieun} & EMNLP'22 & DCG & CLIP-ViT-B/32 & Vanilla Transformer & \textcolor{mix}{MMN/RMN/RNL | MovieUN} \\
    \citet{zhu2022end} & COLING'22 & DCG & CNN & T5 & \textcolor{method}{Seg+Cap} \\
    \citet{deng2021sketch} & CVPR'21 & DCG & CNN & Vanilla Transformer & \textcolor{method}{SRG} \\
    \citet{wang2021end} & ICCV'21 & DCG & Deformable Transformer & Vanilla Transformer + LSTM & \textcolor{method}{PDVC} \\
    \citet{liu2021video} & ACL'21 & DCG & BMN & BERT + Vanilla Transformer & \textcolor{method}{VPCSum} \\
    \citet{zhu2020actbert} & CVPR'20 & DCG & CNN + Faster R-CNN & Tangled Transformer & \textcolor{method}{ActBERT} \\
    \citet{lei2020tvr} & ECCV'20 & DCG & XML & not applicable & \textcolor{mix}{ XML | TVR} \\
    \citet{fang2020video2commonsense} & EMNLP'20 & DCG & CNN + LSTM & Vanilla Transformer & \textcolor{mix}{V2C-Transformer | V2C} \\
    \citet{gurari2020captioning} & ECCV'20 & DCG & not applicable & not applicable & \textcolor{dataset}{ VizWiz-Captions} \\
    \citet{lin2020semi} & EMNLP'20 & DCG & ECO & Vanilla Transformer & \textcolor{method}{SC-SSL} \\
    \citet{shigeto2020video} & LREC'20 & DCG & ResNet-152 + ResNeXt-101 & GRU & \textcolor{dataset}{STAIR Actions} \\
    \citet{huang2020multimodal} & AACL'20 & DCG & Multimodal Transformer & Vanilla Transformer & \textcolor{dataset}{ViTT} \\

    \bottomrule
    \end{tabular}}
    \caption{A collection of studies related to dense video captioning (DVC). We denote works that introduce a new dataset in \textcolor{dataset}{yellow}, works that propose a new method in \textcolor{method}{blue}, and works that deliver both in \textcolor{mix}{green}.}
    \label{tab:dvc-paper-collection}
\end{table*}

This survey is structured as follows: Section~\ref{sec:dvc} provides a brief review of works on DVC; Section~\ref{sec:post-editing} offers an overview of post-editing techniques for AD generation (APE); Section~\ref{sec:eval} discusses the evaluation of AD generation systems (ADE); Section~\ref{sec:future} explores future research directions for developing automatic AD generation systems; and Section~\ref{sec:concl} summarizes the main takeaways of this survey.

\section{Dense Video Captioning for AD Script Generation}

\label{sec:dvc}

Dense video captioning (DVC) addresses the challenge of establishing connections between clips in videos and their natural language descriptions \citep{qasim2023dense}. The term \textit{dense} in DVC signifies the aim to capture as much information as possible to fit the description requirements, which makes DVC a necessary step for AD generation. Typically, DVC outputs multiple sentences as descriptions \citep{liu2021video}.

\begin{figure}[!htb]
    \centering
    \includegraphics[width=\columnwidth]{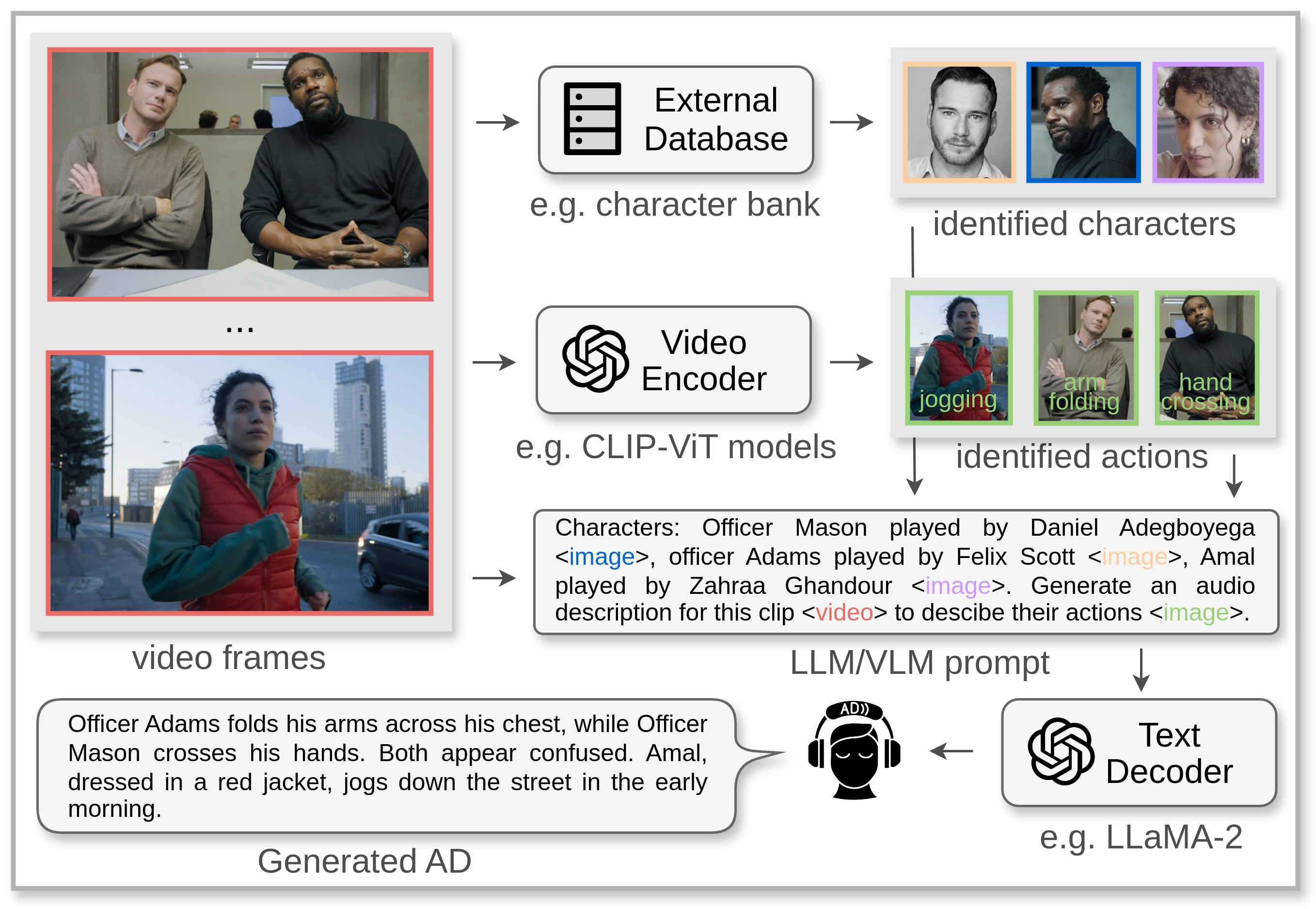}
    \caption{\textbf{Process of DVC}: it is often composed of two sub-tasks, i.e., visual feature extraction (VFE)---where a visual encoder decides \textbf{whom} and \textbf{what} to describe, and dense caption generation (DCG)---where a text decoder works on \textbf{how} to describe. Film taken as example: \textit{Baghdad in My Shadow (2019)}.}
    \label{fig:enter-label}
\end{figure}

For the purpose of automatic AD generation, two sub-tasks of DVC are of particular importance: 
\begin{itemize}
    \item \textbf{Visual Feature Extraction} (VFE), which involves extracting visual features with a visual encoder within videos that are of interest for DVC. When specialized for AD generation, it means identifying characters (\textbf{whom}) and events (\textbf{what}) that are important for ADs.
    \item \textbf{Dense Caption Generation} (DCG), which pertains to the methods of automatically generating ADs in the form of natural language scripts derived from the detected event proposals (\textbf{how}).
\end{itemize}

In this survey, we include works on identifying actions, events, and scenes within the context of DVC, as they are all commonly represented in ADs produced by professionals. While actions refer to specific movements (e.g., eating, running, leaving) performed by a subject, typically classified into predefined classes and extracted as bounding boxes within video frames, events can be understood as a series of actions occurring within a temporal range in the video. Scenes, correspondingly, refer to coherent segments of a video that depict a specific event or sequence of actions happening in a continuous time frame, often within a particular setting.

Since  ADs are typically inserted during silent moments between dialogues to avoid interference with the ongoing narration---a task that is relatively straightforward---this survey does not delve deeply into techniques for identifying specific video frames for AD insertion.
Instead, we focus on reviewing VFE and DCG methodologies to improve AD generation quality, particularly with the integration of LLMs/VLMs.

Next, in Section \ref{sec:epd} and \ref{sec:dcg}, we provide a summary of the relevant VFE and DCG methodologies that can be applied to AD generation. We list relevant studies in Table~\ref{tab:dvc-paper-collection}. 

\subsection{Visual Feature Extraction}
\label{sec:epd}

Convolution-based visual feature extractors \citep{krizhevsky2012imagenet, simonyan2014very, he2016deep} were the mainstream of computer vision research for a long time. In recent years, the Vision Transformer (ViT; \citet{dosovitskiy2020vit}) has emerged as a central component in modern VFE systems and has been integrated into numerous multimodal VLMs such as CLIP \citep{radford2021learning}.
Although not being the first work that tries to apply Transformers for CV tasks, ViT gained its popularity due to its simple design and  scalability.   

ViT preserves the foundational architecture of the standard Transformer by mapping an image into a sequence of patches, analogous to text tokens in NLP tasks. These patches are then processed to produce linear embeddings, which serve as the inputs to the standard Transformer encoder. In comparison to convolutional kernels, the self-attention mechanism in ViT can be viewed as a soft convolutional inductive bias, while being capable of effectively capturing global dependencies within the input patches \citep{d2021convit, raghu2021vision}.
This enables ViT models to exhibit exceptional feature extraction capabilities, resulting in its outstanding performance across various CV tasks \citep{chen2021crossvit, bhojanapalli2021understanding, li2022exploring, minderer2022simple}.

Although ViT-based solutions offer significant advantages, they are often constrained by the high complexity associated with exhaustive self-attention computations. 
To mitigate this challenge, recent research has concentrated on improving efficiency through the development of advanced self-attention computation techniques \citep{huang2021shuffle, liu2021swin, yang2022lite, liu2022video, wu2022pale, hassani2023neighborhood}, dynamic feature selection methods \citep{rao2021dynamicvit, yin2022vit, chen2023efficient}, and optimized memory scheduling strategies \citep{wu2022memvit, liu2023efficientvit, yun2024shvit}. 
These approaches are typically evaluated using video action recognition benchmarks, where the system's output is categorized into predefined action classes. AD generation systems often utilize these identified actions as part of the events (\textbf{what}) that need to be described.

A crucial additional step for VFE in the context of AD generation is the identification of characters involved in events \citep{kukleva2020learning, brown2021face}. This process usually involves comparing the extracted features against stored character profiles in an external database \citep{brown2021automated, han2023autoadii}, which is essentially an information retrieval task. However, creating and indexing large databases for streaming media content is both costly and often impractical due to copyright concerns. Consequently, by utilizing knowledge encoded in pre-trained VLMs, zero-shot character identification has emerged as a more economical and feasible solution \citep{bhat2023face, patricio2023zero, xie2024autoad}.

\subsection{Dense Caption Generation}
\label{sec:dcg}

\begin{figure}
    \centering
    \includegraphics[width=\columnwidth]{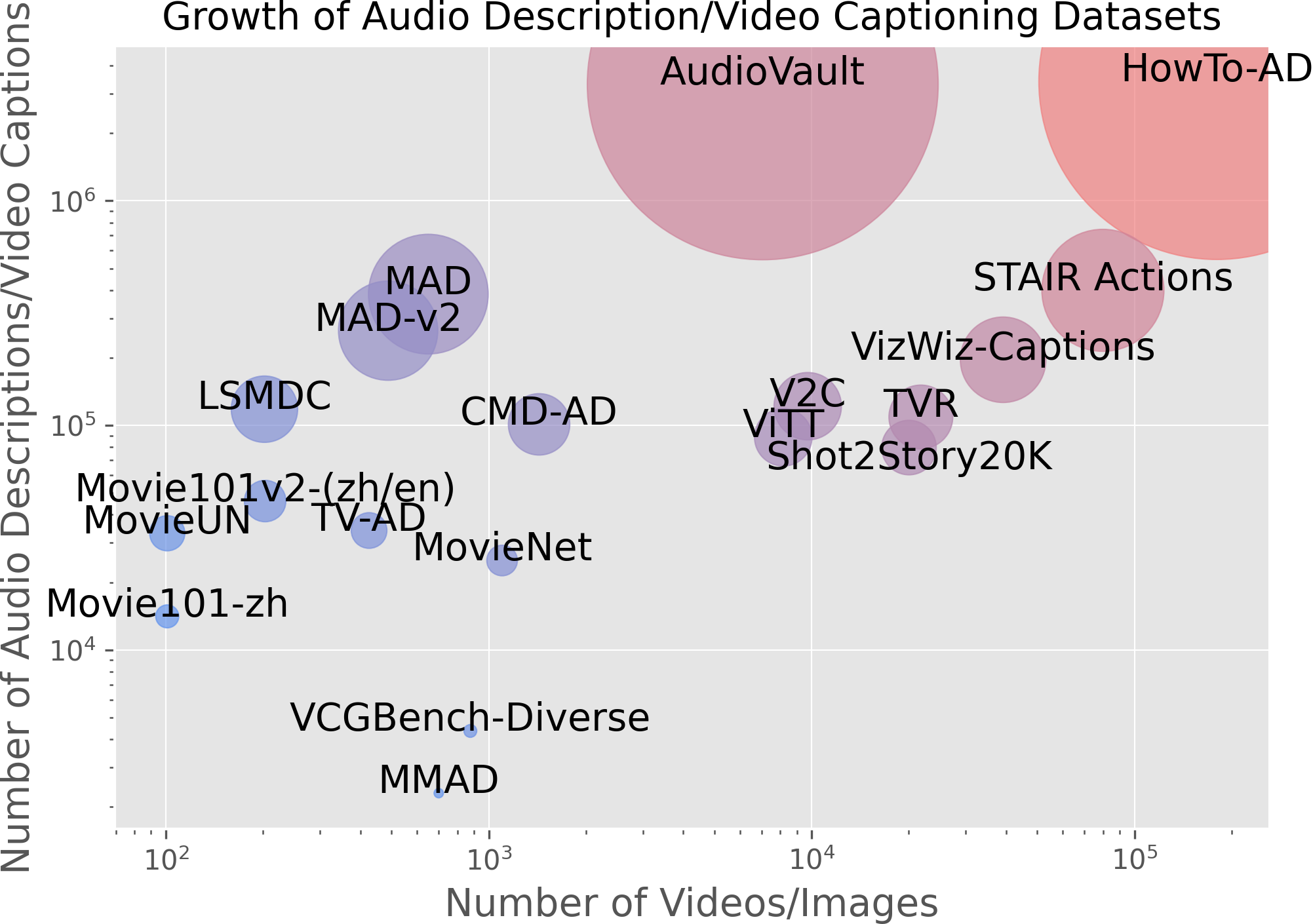}
    \caption{Datasets for AD generation/video captioning. The numbers are visualized in log scale. Red color indicates more recent datasets.}
    \label{fig:dataset-growth}
\end{figure}

To generate dense video captions from extracted visual features, advanced VLMs are employed to learn the alignment between the generated text tokens and the corresponding visual tokens. For this purpose, multimodal DCG datasets are needed for LLMs to learn the alignment. 

Creating large-scale datasets for training is a resource-intensive endeavor. To reduce the workload, researchers often augment existing video datasets with text captions \citep{lei2020tvr, huang2020multimodal, gurari2020captioning, shigeto2020video, huang2020movienet, oncescu2021queryd, yue2023movie101, han2023shot2story20k, yue2024movie101v2}, or retrieve video counterparts for text annotations \citep{rohrbach2017movie, soldan2022mad, zhang2022movieun}. Regardless of the annotation approach, subtitles play a crucial role in creating these video-text alignments, often transcribed using automatic speech recognition (ASR) models such as Whisper-based models \citep{bain2022whisperx, radford2023robust}.

In recent years, DCG research has increasingly focused on zero-shot caption generation \citep{yang2023multicapclip}, contextualizing visual features using separate image and video encoders \citep{maaz2024videogpt+}, developing end-to-end captioning models \citep{zhu2020actbert, deng2021sketch, wang2021end, zhu2022end}, enhancing model efficiency through memory storage \citep{he2024ma, Zhou_2024_CVPR}, and fine-tuning models on well-curated data \citep{luo2024shotluck}. Additionally, efforts have been made to augment ADs with detailed environmental and object information \citep{fang2020video2commonsense, jung2023retrieval, ye2024mmad}. While these efforts have achieved remarkable performance in generating video captions, they have rarely been fully dedicated to the specific task of AD generation.

Generating high-quality ADs requires the integration of both local context (features within the current video frame) and global context (features from past or future frames). The typical length of movies and other streaming media has caused a trade-off between inference speed, which is particularly critical for live video captioning \citep{blanco2024live}, and the quality of the ADs.

To tackle the challenges of curating supervised data and generating high-quality ADs, researchers at the University of Oxford introduced a series of cutting-edge models. In their initial work, \citet{han2023autoad} bridge foundation LLM (GPT) and VLM (CLIP) models to perform vision-conditioned AD generation, optimizing the following loss function:
\begin{align*}
    \mathcal{L}_{NLL} = -\log p_\Theta \left( \mathcal{T}_{\mathbf{x}_i} | \mathbf{h}_{\mathbf{x}_i}, \mathbf{h}_{\mathrm{AD}}, \mathbf{h}_{\mathrm{Sub}} \right),
\end{align*}
where the model leverages representations of context frame ($\mathbf{h}_{\mathbf{x}_i}$ from CLIP with $\mathbf{x}_i$ being the current video clip), subtitles ($\mathbf{h}_{\mathrm{Sub}}$), and previous ADs ($\mathbf{h}_{\mathrm{AD}}$) to enhance the generated AD ($\mathcal{T}_{\mathbf{x}_i}$). Thanks to its modular design, the model can be pre-trained even with limited large-scale data for one modality (i.e., visual-only or text-only pre-training). Their \textit{AutoAD} model demonstrated significant qualitative and quantitative improvements in AD generation.

In their subsequent work \citep{han2023autoadii}, the authors addressed the character naming issue in \textit{AutoAD} by introducing a database containing character names, actor profiles, and CLIP face features. Additionally, the authors explored various methods for predicting AD temporal proposals, specifically identifying movie pauses suitable for AD insertion. With these enhancements, their \textit{AutoAD II} model achieved further improvements in AD generation quality.

Recently, the authors extended their research with the publication of \textit{AutoAD III} \citep{han2024autoadiii}, introducing two new AD datasets created from raw videos with soundtracks, a novel Q-Former-based architecture for AD generation, and two new AD evaluation metrics. This work underscores the advancement of LLM/VLM participation in AD generation. 

In their latest work \citep{xie2024autoad}, the authors explore a two-stage zero-shot approach to AD generation. Initially, a VLM is prompted with key information, such as character identities and their interactions, to generate dense captions. These captions are then further summarized into ADs by an LLM. The authors evaluated their \textit{AutoAD-Zero} model on a custom dataset, \textit{TV-AD}, achieving competitive results even when compared to supervised models trained on gold-standard ADs. 

\textit{AutoAD} papers illustrate the effectiveness of utilizing LLMs and VLMs for AD generation. Recently, prompt-based pipelines employing GPT-4V as the video encoder and GPT-4 as the text decoder have shown significant potential in producing ADs that align with human production standards \citep{lin2023mm, chu2024llm}. However, further enhancements in generation quality may require the integration of expert knowledge to achieve more coherent and contextually accurate AD narrations.

\section{AD Post-editing}
\label{sec:post-editing}

%\subsection{AD Script Editors}

While a simple video player and a Word editor may be enough for audio describers to edit AD scripts \citep{minutella2022audio}, a variety of specialized professional AD software is available to enhance the quality and efficiency of this process. These tools include options such as \citet{captioningstar2006}, VDManager \citep{gagnon2010computer}, LiveDescribe \citep{branje2012livedescribe}, \citet{youdescribe2013}, \citet{3playmedia2007}, \citet{livevoice2016}, \citet{atotalaccess2020}, Rescribe \citep{pavel2020rescribe}, \citet{frazier2021}, \citet{stellar2022}, and \citet{audiblesight2023}. These platforms offer advanced features tailored specifically for creating, editing, and managing human- or machine-generated ADs, thus providing significant advantages over more general-purpose tools.

%Recently, automatic AD generation software has begun to emerge in the market of accessibility products, signaling a significant advancement in the field \citep{audiblesight2023, appypie_audiodesc_2023, verbit_audiodesc_2023, amberscript_audiodesc_2023}. 
%However, while AI plays a substantial role in generating ADs, its involvement in post-editing processes remains limited, with the level of automation in this area still being relatively low.

Machine-generated ADs often contain grammatical errors and other undesirable elements. To address this issue, text editing models are developed and trained to improve the quality of these texts. These models typically utilize training data that includes human-simplified or corrected texts \citep{faltings2021text, kim2022improving, zhang2023non}. Among these, many LLM-based models are fine-tuned with instructions \citep{raheja2023coedit, raheja2024medit, shu2024rewritelm, ki2024guiding}, while others are trained using semi-autoregressive or non-autoregressive decoding techniques \citep{mallinson2022edit5, agrawal2022imitation, zhang2023non}.

Currently, post-editing is still crucial for ensuring adherence to AD production principles (e.g., Figure~\ref{fig:pipeline} (c)). However, we contend that future research should focus on the automation of AD generation, thereby eliminating the need for human post-editing.

\section{AD Evaluation}
\label{sec:eval}

\subsection{Automatic Evaluation}

Automatic evaluation of ADs typically involves comparing the generated ADs to the gold standards. Classic text generation metrics are employed to assess: 1) textual relevance through N-gram overlaps (e.g., BLEU \citep{papineni2002bleu}, ROUGE-L \citep{lin2004rouge}, METEOR \citep{banerjee2005meteor}, and CHRF \citep{popovic2015chrf}); or 2) embedding-based semantic similarity between the generated and ground-truth ADs (e.g., MoverScore \citep{zhao2019moverscore}, BERTScore \citep{zhang2020bertscore}, BARTScore \citep{NEURIPS2021_e4d2b6e6}, (Ref)CLIPScore \citep{hessel2021clipscore}, EMScore \citep{shi2022emscore}, and (Ref)PAC-S \citep{sarto2023positive}).

Given the multimodal nature of AD generation, image and video captioning metrics are also widely employed to evaluate the quality of generated ADs. Unlike traditional text generation metrics, CIDEr \citep{vedantam2015cider} assesses N-gram overlaps between generated captions and a set of reference captions, under the assumption that effective machine-generated captions should resemble those produced by a diverse group of humans. SPICE \citep{anderson2016spice} evaluates captions by converting them into graph structures and comparing their semantic propositions. SPIDEr \citep{liu2017improved}, a linear combination of SPICE and CIDEr, measures both semantic accuracy and syntactic fluency in generated captions. 
\citet{fujita2020soda} introduced SODA, a metric designed to evaluate generated captions based on their effectiveness in describing the video story, with particular emphasis on maintaining temporal order and textual coherence of the captions.
BERTHA \citep{lebron2022bertha}, a BERT-based model trained on human-evaluated captions, is designed to maximize the correlation between automatic evaluation and human judgment.
Recently, new image captioning metrics based on multimodal language models have been introduced to enhance scoring explainability \citep{hu2023infometic, chan2023clair, lee2024fleur}, further highlighting the growing role of LLMs and VLMs in evaluating machine-generated ADs.

Specialized AD evaluation metrics have also been explored. \citet{yue2023movie101} proposed MNScore, which evaluates AD quality by accounting for both semantic similarity and character name generation. \citet{han2024autoadiii} introduced two additional metrics: CRITIC, a coreference-based approach for measuring character recognition, and LLM-AD-Eval, a metric based on LLM-prompting that assesses the overall quality of ADs.

\subsection{Human Evaluation}

Many human evaluation works focus on how AD end users in different countries perceive ADs in terms of their \textbf{usefulness}.  

\citet{lopez2018audio} explored the usefulness of film and television ADs in the UK. The authors noted that while ADs are useful, there is still room for improvement, particularly in terms of personalization and the integration of sound design techniques (which are proven to be effective in their later work \citep{lopez2021enhancing}), which could potentially create a more immersive experience for AD end users.
\citet{reviers2018studying} analyzed Dutch films conducted in Flanders and the Netherlands and confirmed the found of idiosyncratic language patterns.
\citet{ferziger2020audio} examined the reception of ADs in cultural events in Israel, such as theater patrons, where participants reported high levels of overall satisfaction with the AD services provided. 
\citet{bausells2022audio} explored the reception of ADs in a pedagogical setting for foreign language (Spanish) teaching. Their study found that, depending on the students' perceived level of difficulty, ADs can be highly helpful in developing transferable and communicative skills such as summarizing and narrating.
\citet{arias2023audio} found that AD scripts in Catalan adhere to characteristics of ``easy-to-understand'' language, utilizing simple syntax and lexicon.
\citet{yang2023pilot} conducted a systematic study on the availability and reception of AD services in mainland China, revealing that, despite significant challenges such as a shortage of AD professionals, limited foundational research, and copyright constraints, AD end users expressed satisfaction with the quality of services even though their comprehension of the movies remained low.
\citet{leong2023audio} investigated the effectiveness of ADs in aiding blind and visually impaired individuals to interact within 3D virtual environments. The authors concluded that ADs alone are insufficient for facilitating navigation and orientation in such environments and recommended the integration of additional auditory cues such as sound landmarks.

Other relevant studies focus on evaluating the nature of ADs themselves, rather than their functions and effects.
For example, \citet{jekat2018reception} compare the reception of two distinct AD styles: \textbf{descriptive} and \textbf{interpretative}, among German-speaking AD end users. The study found that, contrary to expectations, users reported a more immersive movie experience with interpretative ADs. This finding challenges the traditional preference for descriptive ADs, which have long been the standard among many German public broadcasters.  
\citet{gallego2020defining} investigated the extent to which subjective ADs in art museums are preferred by blind and visually impaired individuals. Their methodology, which integrates cognitive linguistics and art theory, offers valuable insights into how subjective ADs can effectively enhance guided tours in art museum settings.
By contrast, \citet{munoz2023multimodal} focused on analyzing objectivity in the ADs of Spanish Netflix videos. The results indicate that these ADs are neither purely objective nor entirely subjective.
\citet{wang2022evaluating} proposed six distinct methods for assessing the emotional responses of AD users during museum tours.

Human evaluation of ADs often spans multiple research domains, including psychology, pedagogy, and cognitive science, employing methodologies that range from traditional questionnaire-based approaches to measuring neural activities such as EEG signals. However, these interdisciplinary insights have yet to be integrated into the AD generation process using LLM and VLM models. We therefore advocate for multidisciplinary collaborations between AI and non-AI communities to jointly address the challenges in AD generation, reduce technical barriers, and adhere more closely to user-centered principles. 

\section{Future Research}
\label{sec:future} 

AD generation is a complex task that extends far beyond the mere application of LLMs and VLMs. Building on the research reviewed above, we outline the following future research directions.

\subsection{AD Generation with Human Preferences}

Although general international AD standards, such as ISO/IEC TS 20071-21:2015\footnote{\url{https://www.iso.org/standard/63061.html}}, have already existed for a long time, individual nations and audio describers often follow their own inclusive guidelines for AD production \citep{mazur2024same}. These specific rules have not been incorporated into the tuning process of LLMs/VLMs. Consequently, tuning AD generation systems with these human-crafted guidelines would be beneficial. This could be achieved through LLM alignment techniques \citep{ouyang2022training, rafailov2024direct, meng2024simpo, ethayarajh2024kto}, where AD generation models are optimized to produce outputs that align with human preferences. 

\subsection{Personalized AD Generation}

Recent research indicates that varying degrees of visual impairment can significantly influence perception of ADs \citep{seve2024studie}, which underscores the importance of personalizing AD generation according to individual requirements of end users \citep{ natalie2024audio}.
In addition, AD generation systems for movies and TV episodes should differ from those for art galleries and museums. Moreover, AD generation systems tailored for individuals without intellectual disabilities should be distinct from those intended for persons with intellectual disabilities. Combining AD generation with text simplification could further enhance accessibility for diverse audiences \citep{braun2021innovation}. 

Last but not least, AD generation systems should also prioritize scenarios such as higher education, where ADs are crucial in supporting blind students and students with visual impairments for better learning experience. 

\subsection{Machine Translation of AD Scripts}

Given that ADs are often available in only one language, research has focused on utilizing machine translation models to translate ADs from one language to another \citep{matamala2016accessibility}. This approach aims to facilitate the production of ADs in situations where multilingual audio describers are not available. \citet{fernandez2016machine} tested machine translation models on English-Catalan AD script pairs, while \citet{vercauteren2021evaluating} conducted similar research with AD script pairs of English-Dutch. 
\citet{matamala2016building} built a multilingual multimodal corpus for ADs.
\citet{torne2016machine} presented an evaluation of five English-Catalan AD translation systems, employing both automatic and subjective post-editing metrics to assess their performance.
These studies not only confirmed the potential of machine translation models for AD translation but also highlighted the significant human post-editing efforts required to achieve satisfactory translation quality.

%\citet{fischer2024swissadt} presented \textit{SwissADT}, the first multilingual and multimodal AD translation system designed specifically for translating AD scripts in Switzerland's three main languages by utilizing LLMs and incorporating visual inputs from video clips. Their system uses data collected from Swiss national television and synthetic ADs generated with DeepL, demonstrating improved translation quality through both automatic and human evaluations.

However, while promising strides have been made in AD translation research, it remains under-explored and not yet fully integrated into AD production pipelines. More research is needed to refine these models and establish their role in practical applications, ensuring they meet the high standards required for AD production.
\section{Conclusions}
\label{sec:concl}

As an inclusive product, ADs have greatly enhanced access to information for blind persons and persons with visual impairments. However, traditional AD production, which relies on human audio describers, is often both costly and time-consuming. In contrast, generative AI technologies, such as LLMs and VLMs, have shown significant potential in automating the AD generation process. In this survey, we reviewed the technologies that are applicable to AD generation, including dense video captioning, (automatic) post-editing, and AD evaluation. As emphasized by \citet{hirvonen2023co}, AD production should adhere to user-centric principles, and we believe that LLMs and VLMs can play a crucial role in supporting this requirement.

\section{Limitations}
\label{sec:limit}

Our study has two main limitations: 1) We did not explore other DVC sub-tasks, such as video temporal grounding, which involves associating a natural language query with a specific temporal video segment. This omission is because ADs are meant to serve as the final output of AD generation systems, not as queries for retrieving content in videos that blind individuals or individuals with visual impairment cannot perceive. However, we acknowledge that blind individuals and individuals with visual impairment may have information retrieval needs, such as revisiting previous clips in a video, potentially using voice commands. Unfortunately, we found no relevant literature addressing this problem. 2) Given that generated ADs are typically inserted during silent moments between dialogues to avoid interfering with the ongoing narration, this survey does not thoroughly examine techniques for identifying suitable pauses for AD insertion.

%\section*{Acknowledgments}

\bibliography{custom}

% \appendix

% \section{Example Appendix}
% \label{sec:appendix}

% This is an appendix.

\end{document}